\documentclass[10pt,twocolumn,letterpaper]{article}

\usepackage{iccv}
\usepackage{times}
\usepackage{epsfig}
\usepackage{graphicx}
\usepackage{amsmath}
\usepackage{amssymb}

\usepackage{enumitem}
\usepackage{booktabs}
\usepackage{multirow}
\usepackage{graphicx}
\usepackage{caption}
\usepackage{subcaption}
\usepackage[ruled,linesnumbered]{algorithm2e}
\usepackage{rotating}
\setlist[itemize]{leftmargin=4mm}
\usepackage{dblfloatfix} 

\usepackage[symbol*]{footmisc}
\usepackage{pifont}
\usepackage{authblk}
\usepackage[export]{adjustbox}

\newcommand{\cmark}{\ding{61}}%
\newcommand{\xmark}{*}%
\newcommand{\zmark}{\ding{67}}%

\usepackage[breaklinks=true,bookmarks=false]{hyperref}

\iccvfinalcopy 

\def\iccvPaperID{****} 

\graphicspath {{figures/}}

\usepackage[dvipsnames]{xcolor}

\newcommand{\wadim}[1]{\textcolor{blue}{#1}}

\newcommand{\comment}[1]{}

\ificcvfinal\fi

\begin{document}

\title{DeceptionNet: Network-Driven Domain Randomization}

\author[ \xmark,\cmark]{Sergey Zakharov}
\author[ \zmark]{Wadim Kehl}
\author[ \xmark,\cmark]{Slobodan Ilic}
\affil[\xmark]{ Technical University of Munich\quad \zmark~Toyota Research Institute\quad \cmark~Siemens Corporate Technology
{\normalsize \tt{sergey.zakharov@tum.de}}, {\normalsize \tt{\, wadim.kehl@tri.global}}, {\normalsize \tt{\, slobodan.ilic@siemens.com}}}

\maketitle
\ificcvfinal\thispagestyle{empty}\fi

\begin{abstract}
    We present a novel approach to tackle domain adaptation between synthetic and real data. Instead, of employing "blind" domain randomization, i.e., augmenting synthetic renderings with random backgrounds or changing illumination and colorization, we leverage the task network as its own adversarial guide toward useful augmentations that maximize the uncertainty of the output. To this end, we design a min-max optimization scheme where a given task competes against a special deception network to minimize the task error subject to the specific constraints enforced by the deceiver. The deception network samples from a family of differentiable pixel-level perturbations and exploits the task architecture to find the most destructive augmentations. Unlike GAN-based approaches that require unlabeled data from the target domain, our method achieves robust mappings that scale well to multiple target distributions from source data alone. We apply our framework to the tasks of digit recognition on enhanced MNIST variants, classification and object pose estimation on the Cropped LineMOD dataset as well as semantic segmentation on the Cityscapes dataset and compare it to a number of domain adaptation approaches, thereby demonstrating similar results with superior generalization capabilities.

\end{abstract}


\section{Introduction}

\begin{figure*}[!t]
	\centering
	\includegraphics[width=1\linewidth]{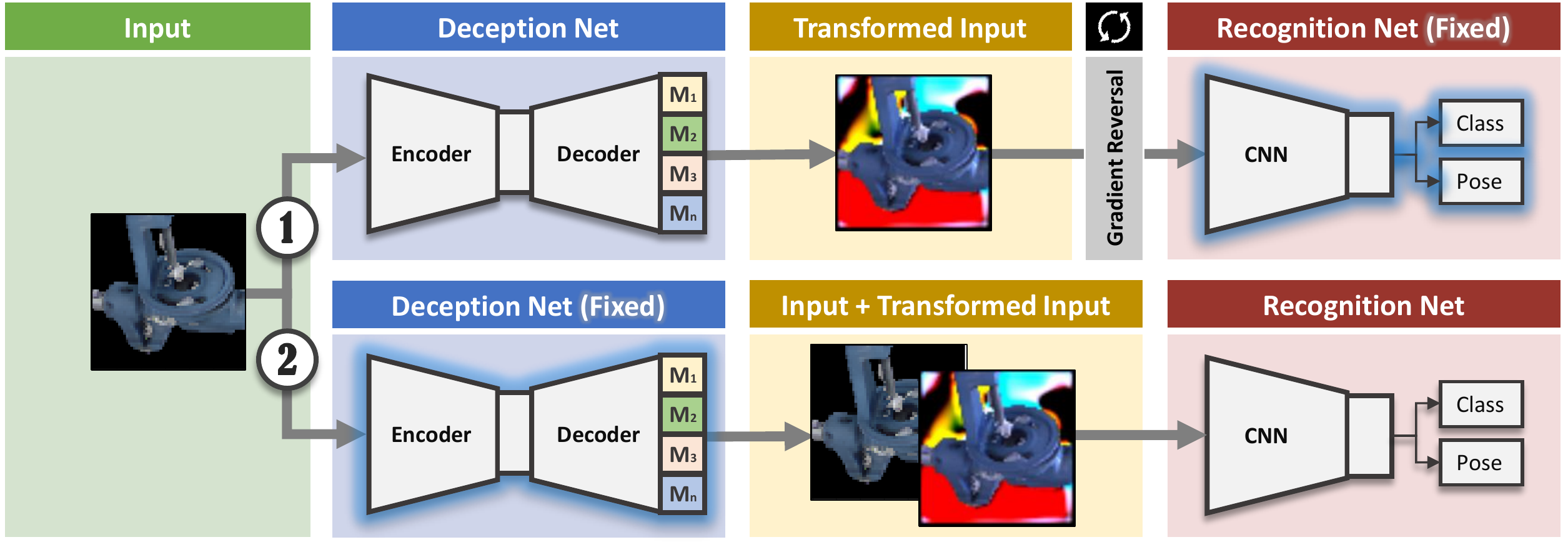}
	\caption{\textbf{Training pipeline.} Training is performed in two alternating phases. \textbf{Phase 1}: The weights of the deception network are updated, while those of the recognition network are frozen. The recognition network's objective is maximized instead of being minimized, forcing the deception network to produce increasingly confusing images. \textbf{Phase 2}: The generated deceptive images provided by the deception network, whose weights are now frozen, are passed to the recognition network and its weights are updated such that the loss is minimized. As a result of this min-max optimization, the input images are automatically altered by the deception network, forcing the recognition network to be robust to these domain changes.}
	\label{fig:training}  
\end{figure*}

The alluring possibility of training machine learning models on purely synthetic data allows for a theoretically infinite supply of both input data samples and associated label information. Unfortunately, for computer vision applications, the domain gap  between synthetic renderings and real-world imagery poses serious challenges for generalization. 
Despite the apparent visual similarity, synthetic images structurally differ from real camera sensor data. First, synthetic image formation produces clear edges with approximate physical shading and illumination, whereas real images undergo many types of noise, such as optical aberrations, Bayer demosaicing, or compression artifacts. Second, the visual differences between synthetic CAD models and their actual physical counterparts can be quite significant. 
Apart from the visual gap, supervised approaches also require cumbersome and error-prone human labeling of real training data in the form of 2D bounding boxes, segmentation masks, or 6D poses \cite{lin2014microsoft, hodan2017t}. For other approaches, such as robotic control learning, solutions must be found by exploration in tight simulation-based feedback loops that require synthetic rendering \cite{Mahler2017, sadeghi2017cadrl, OpenAI2018}. 

The gap between the visual domains is nowadays mainly bridged with adaptation and/or randomization techniques. 
In the case of supervised domain adaptation approaches~\cite{ZeilerF13,Oquab14,BabenkoSCL14, Motiian2017}, a certain amount of labeled data from the target domain exists, while in unsupervised approaches~\cite{ganin2016domain, taigman2016unsupervised,shrivastava2016learning,bousmalis_unsupervised} the target data are available but unlabeled.
In both cases, the goal is to match the source and target distributions by finding either a direct mapping, a common latent space, or through regularization of task networks trained on the source data.
Recent unsupervised approaches are mostly based on generalized adversarial networks (GANs)~\cite{bousmalis_unsupervised, LedigTHCATTWS16,Liu16, taigman2016unsupervised, tzeng2017adversarial, lee2018spigan, sixt2018rendergan, ratner2017learning, antoniou2017data} 
and although these methods perform proper target domain transfers, they can overfit to the chosen target domain and exhibit a decline in performance for unfamiliar out-of-distribution samples. 

Domain randomization methods ~\cite{tobin2017domain,Kehl2017,Manhardt2018,zakharov2018keep,sundermeyerImplicit3DOrientation2018} have no access to any target domain and employ the rather simple technique of randomly perturbing (synthetic) source data during training to make the tasks networks robust to perceptual differences. This approach can be effective, but is generally unguided, and needs an exhaustive evaluation to find meaningful augmentations that increase the target domain performance. 
Last but not least, results from pixel-level adversarial attacks \cite{brown2017adversarial, Su2019} suggest the existence of architecture-dependent effects that cannot be addressed by "blind" domain randomization for robust transfer.

We propose herein a general framework that performs guided randomization with the help of an auxiliary deception network trained in a similar min-max fashion as GAN networks. This is done in two alternating phases, as illustrated in Fig.~\ref{fig:training}. In the first phase, the synthetic input is fed to our deception network responsible for producing augmented images that are then passed to a recognition network to compute the final task-specific loss with provided labels. Then, instead of minimizing the loss, we maximize it via gradient reversal \cite{ganin2015unsupervised} and only back-propagate an update to the deception network parameters. The deception network parameters are steering a set of differentiable modules $M_1,...,M_N$, from which augmentations are sampled. In the next phase, we feed the augmented images to the recognition network together with the original images to minimize the task-specific loss and update the recognition network. In this way, the deception network is encouraged to produce domain randomization by confusing the recognition network, and the recognition network is made resilient to such random changes. By adding different modules and constraints we can influence how much and which parts of the image the deception network alters. In this way, our method outputs images completely independent from the target domain and therefore generalizes much better to new unseen domains than related approaches. In summary, our contributions are:
\begin{itemize}[noitemsep]
    \item DeceptionNet framework that performs a min-max optimization for guided domain randomization;
    \item Various pixel-level perturbation modules employed in such a framework suited for synthetic data;
    \item Novel sequences: MNIST-COCO and Extended Cropped LineMOD that allow to demonstrate our strong generalization capabilities to unseen domains.  
\end{itemize}
In the experimental section we will show that steered randomization by leveraging the network structure actually generalizes much better to new domains than unsupervised approaches with access to the target data while performing comparably well to them on known target domains.

\section{Related Work}
\label{sec:rw}

\paragraph{Domain Adaptation.} Various domain adaptation works put their efforts to bridge the gap between the domains mostly based on unsupervised conditional generative adversarial networks (GANs)~\cite{taigman2016unsupervised,shrivastava2016learning,bousmalis_unsupervised, antoniou2017data} or style-transfer solutions~\cite{gatys2016image}. These methods use an unlabeled subset of target data to improve the synthetic data performance. For example, the authors of~\cite{bousmalis_unsupervised, antoniou2017data} proposed to use GANs to learn the mapping from synthetic images to real. Extending this idea, approaches of \cite{sixt2018rendergan, ratner2017learning} use GANs to tune the parameters of user-defined transformations to fit to the target distribution.  As opposed to GANs, work \cite{devries2017dataset} used a sequence autoencoder to extract the feature vector pairs from the available data, which are then decoded to generate new data samples. 

Alternatively, domain-invariant features that work well for both real and synthetic domains can be learned. ~\cite{Rad2018} mapped real image features to the feature space of synthetic images and used the mapped information as an input to a task-specific network, trained on synthetic data only.

Another example is \textit{DSN}~\cite{bousmalis2016domain}, which proposes the extraction of image representations that are partitioned into two subspaces: private to each domain and one which is shared across domains (learning domain-invariant features). The shared subspace is then used to train a classifier that performs well on both domains. Similarly, \textit{DRIT}~\cite{lee2018diverse} embeds images on a domain-invariant content space (capturing shared information across domains) and a domain-specific attribute space by introducing a cross-cycle consistency loss based on disentangled representations.
Other approaches, such as \textit{DANN}~\cite{ganin2016domain} or \textit{ADDA}~\cite{tzeng2017adversarial} instead focus on adapting the recognition methods themselves to make them more robust to the domain changes. 

\paragraph{Domain Randomization.} However, what if one does not have real data available? The answer for this case is domain randomization. Domain randomization is a popular approach \cite{tobin2017domain,Kehl2017,zakharov2018keep,planche2018seeing, sundermeyerImplicit3DOrientation2018, sadeghi2016cad2rl} that aims to randomize parts of the domain that we do not want our algorithm to be sensitive to. For example, 
~\cite{tobin2017domain} and 
~\cite{sadeghi2016cad2rl} trained complex recognition methods by means of adding variability to the input render data, i.e., different illumination conditions, texture changes, scene decomposition, \etc. This sort of parameterization allows to learn features that are invariant to the particular properties of the domain. The authors of \cite{zakharov2018keep} used a sophisticated depth augmentation pipeline trying to cover possible artifacts of the common commodity depth sensors. It was then used to train a network removing these artifacts from the input and generating a clean, synthetically-looking image. Building on top of this idea, the methods of \cite{planche2018seeing,sundermeyerImplicit3DOrientation2018} extended this to the RGB domain. 

Nevertheless, the main question remains unsolved: What is the main cause of confusion given the domain change? Domain randomization tries to target all possible scenarios, but we do not really know which of them are actually useful to bridge the domain gap. Moreover, covering all possible variations present in the real world by applying simple augmentations is almost impossible.

Our approach can be placed between domain randomization and GAN methods, however, instead of forcing randomization without any clear guidance on its usefulness, we propose to delegate this to a neural network, which we call \textit{deception network}, which tries to alter the images in an automated fashion, such that the task network is maximally confused. Moreover, to do so, we do not require any images, labeled or unlabeled, from the target domain.

\section{Methodology}
\label{sec:mth}

 \begin{figure*}[h]
\centering
        \begin{subfigure}[t]{0.495\textwidth}
                \caption{Deception Modules for \textbf{MNIST}}
                \label{fig:deception_mnist}
                \centering
                \includegraphics[width=\linewidth]{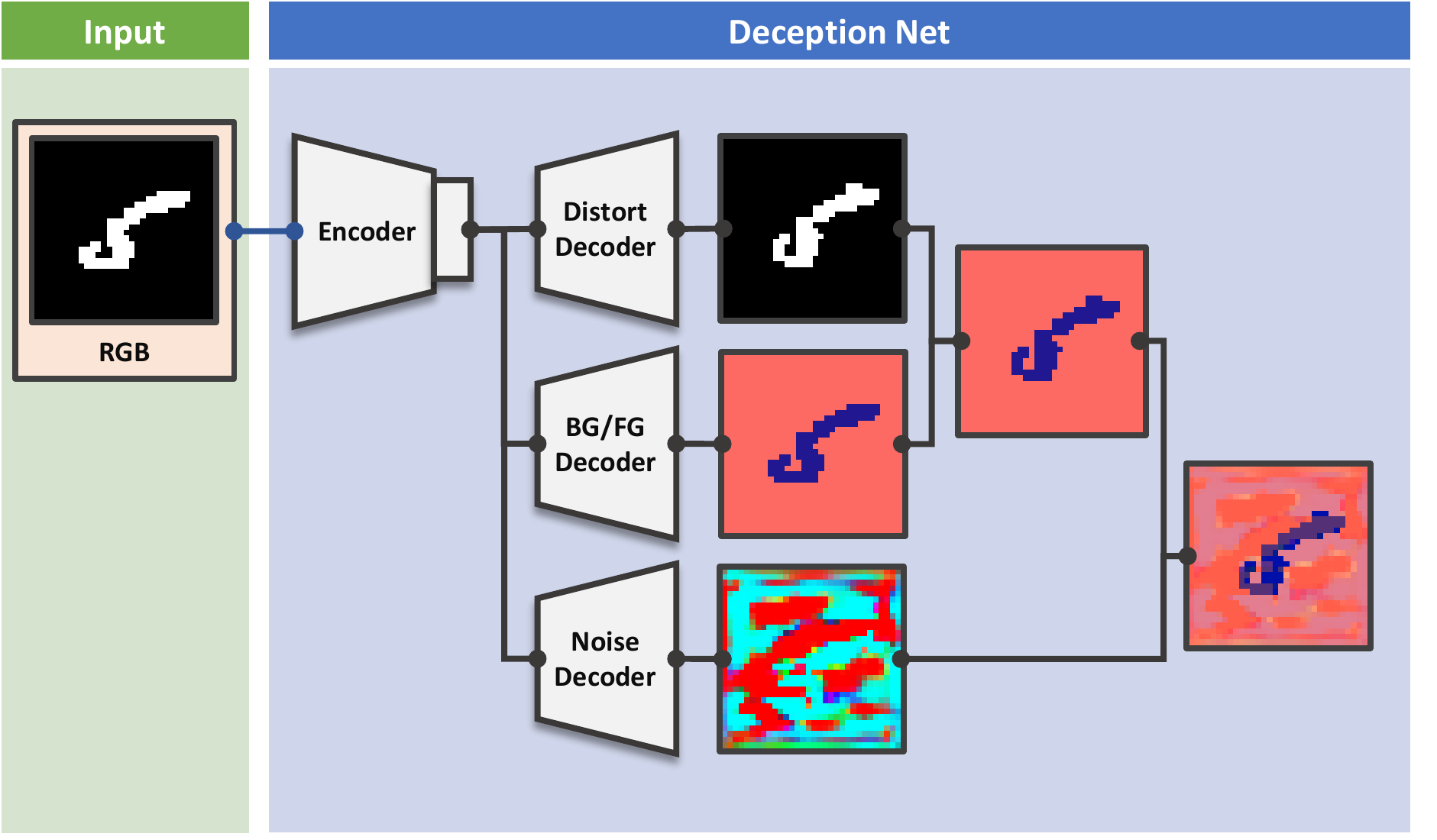}

        \end{subfigure}\hfill
        \begin{subfigure}[t]{0.50\textwidth}
                \caption{Deception Modules for \textbf{LineMOD}}
                \label{fig:deception_lm}
                \centering
                \includegraphics[width=\linewidth]{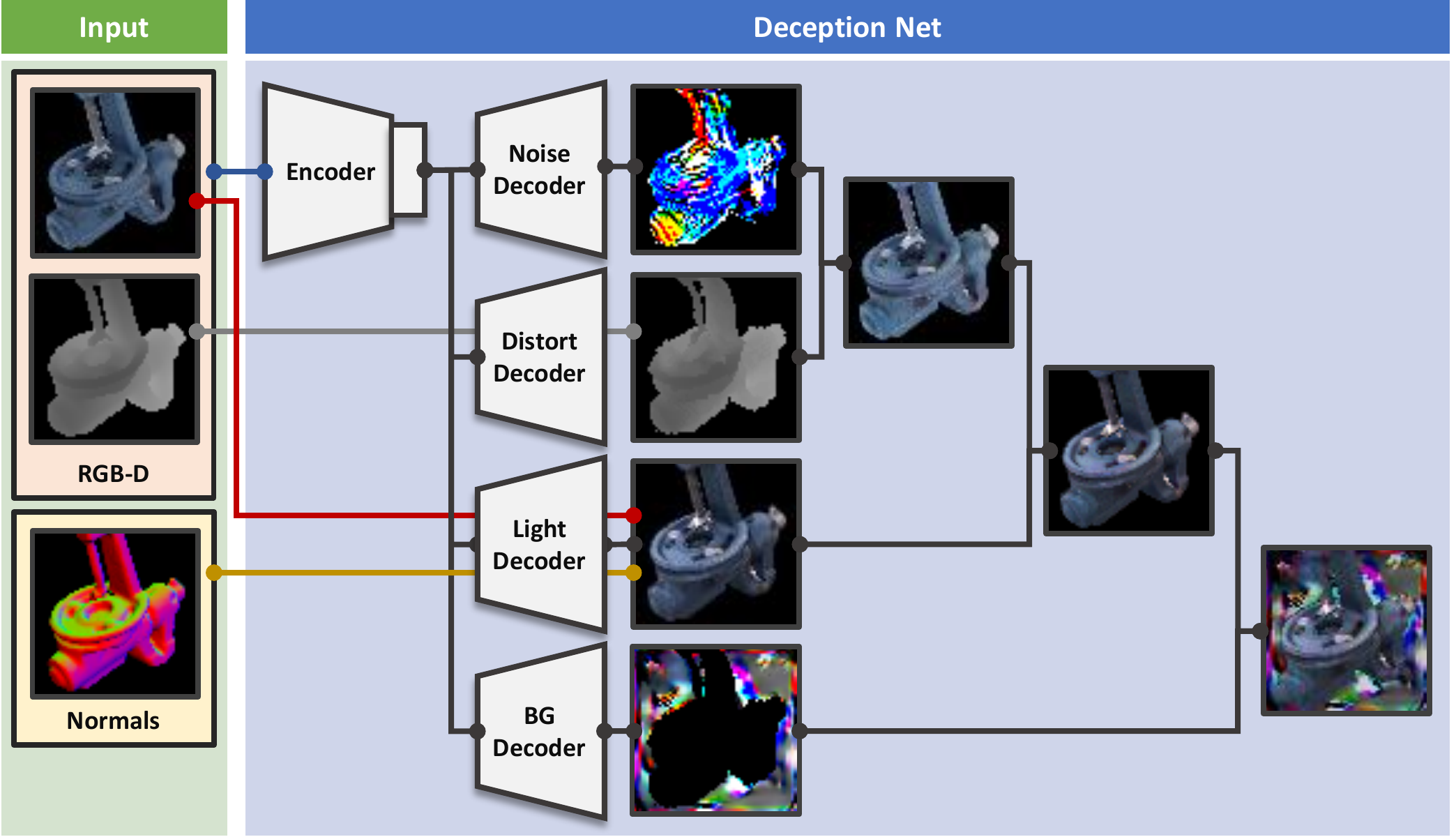}
                
        \end{subfigure}
        \caption{\textbf{Architecture of the deception networks used for the presented experiments.} For the case of \textbf{MNIST} classification, three deception modules are used: the \textit{distortion module} applying elastic deformations on the image, the \textit{BG/FG module} responsible for generating background and foreground colors, and the \textit{noise module} additionally distorting the image by applying slight noise. The \textbf{LineMOD} dataset requires a more sophisticated treatment and features four deception modules: noise and distortion (applied on depth channel only), modules similar to the previous case, pixel-wise BG module and light module generating different illumination conditions based on the Phong model.}\label{fig:decoders}
\end{figure*}

As outlined, our approach towards steered domain randomization is essentially an extension of the task algorithm. Therefore, we have the actual task network \hbox{$T(\mathbf{x}\ ; \mathbf{\theta}_T)\to \hat{\mathbf{y}}$}, which, given an input image $\mathbf{x}$, returns an estimated label $\hat{\mathbf{y}}$ (\eg, class, pose, segmentation mask, \etc), and (2) a deception network $D$ that takes the source image $\mathbf{x^s}$ and returns the deceptive image $\mathbf{x^d}$, which, when provided to the task net $T(D(\mathbf{x^s})) \rightarrow \hat{\mathbf{y}}^d$, maximizes the difference between $\hat{\mathbf{y}}^d$ and $\mathbf{y^s}$.  While the recognition network architectures are standard and follow related work \cite{bousmalis_unsupervised, ganin2015unsupervised}, we will first focus herein on our structured deception network first, and then describe the optimization objective and the training. 

To formalize our pipeline, let $X^s_c:= \mathbf{x^s}_{c,i} \enspace \forall i \in N^s_c$ be a source dataset composed of $N^s_c$ source images $\mathbf{x}^s_c$ for an object of class $\mathbf{c}$. Then, $X^s:=X^s_c \enspace \forall \mathbf{c} \in C$ is the source dataset covering all object classes $C$. A dataset of real images $X^r$ (not used by us for training) is similarly defined.

\subsection{Deception Modules}
The deception network $D$ follows the encoder-decoder architecture where input $\mathbf{x^s}$ is encoded to a lower-dimensional 2D latent space vector $\mathbf{z}$ and given as an input to multiple decoding modules $M_1,..., M_n$. The final output of $D$ is then a weighted sum of decoded outputs $\mathbf{x^d} := \sum_i \mathbf{w_i} \cdot M_i(\mathbf{z})$ where $\mathbf{w_i} \in [0, 1]^{H \times W}$ act as spatial masking operations. While such a formulation allows for flexibility, the decoders must follow a set of predefined constraints to create meaningful outputs and leverage an inherent image structure instead of finding trivial mappings to decrease the task performance (e.g., by decoding always to 0). Note that our proposed framework is general and, thus, requires instantiations of the deception network for specific datasets. Similar to architecture search, discovering the "best" instantiation is infeasible, but good ones can be found by analyzing the data source. After a reasonable experimentation we settled on certain configurations for MNIST (RGB) and LineMOD (RGB-D), depicted in Fig. \ref{fig:decoders}. We continue by providing more detail on the used decoder modules and their constraint ranges.

\subsubsection{Background Module ($\textsc{BG}$)}
Since our source images have black backgrounds, they hardly transfer over to the real world with infinite background variety, resulting in a significant accuracy drop. \cite{Kehl2017, Manhardt2018} tackle this problem by rendering objects on top of images from large-scale datasets (e.g., MS COCO~\cite{lin2014microsoft}).

Instead, our background module produces its output by chaining multiple upsampling and convolution operations. While the output is rather simple at start, the module regresses very complex and visually confusing structures in the advanced stages of training. 

For MNIST, we used a simpler variant that outputs a single RGB background color $\in [0, 1]$ and an RGB foreground bias $\in [0.1, 0.9]$ (restricted not to intersect with the background color). To form the output, we first apply the background color and then add the foreground bias using the mask. We ensure that the final values are in the range $[0, 1]$.

\subsubsection{Distortion Module ($\textsc{DS}$)}

The module is based on the idea of the elastic distortions first presented in \cite{simard2003best}. Essentially, a 2D deformation field is randomly initialized from $[-1, 1]$ and then convolved with a Gaussian filter of standard deviation $\sigma$. For large values of $\sigma$, the resulting field approaches 0, whereas smaller values of $\sigma$ keep the field mostly random. However, the moderate values of $\sigma$ make the resulting field perform elastic deformations, where $\sigma$ defines the elasticity coefficient. The resulting field is then multiplied by a scaling factor $\alpha$, which controls the deformation intensity.

Our implementation closely follows the described approach but we use the decoder output as the distortion field and apply resampling, similar to spatial transformer networks~\cite{jaderberg2015spatial}. We fix $\sigma=4$, but learn both $\alpha \in (0,5]$ and the general decoder parameters. This means that the network itself controls where and how much to deform the object. 

\subsubsection{Noise Module ($\textsc{NS}$)}
Applying slight random noise augmentation to the network input during training is common practice. In a similar fashion, we use the noise decoder to add generated values to the input. The noise decoder regresses a tensor of the input size with values in the range $[-0.01, 0.01]$, which are then added to the input of the module.

\subsubsection{Light Module ($\textsc{L}$)}
Another feature not well covered by synthetic data is proper illumination. 
Recent methods \cite{Kehl2017, Manhardt2018, Hinterstoisser2017, zakharov2019dpod} prerender a number of synthetic images featuring different light conditions. Here, we instead implement differentiable lighting based on the simple Phong model \cite{Phong75}, which is fully operated by the network. While more complex parametric and differentiable illumination models do exist, we found this basic approach to already work quite well.

The module requires surface information which is provided in form of normal maps. From this, we generate three different types of illumination, namely ambient, diffusive, and specular. The light decoder outputs a block of 9 parameters that are used to define the final light properties, i.e., a 3D light direction, an RGB light color (restricted to the range of $[0.8, 1]$), and a weight for each of the three illumination types ($\mathbf{w_a} \in [0.6, 1], \mathbf{w_d} \in [0, 1], \mathbf{w_s} \in [0, 1]$).

\begin{figure*}[t]
\centering
        \begin{subfigure}[b]{0.19\textwidth}
                \caption{MNIST}
                \label{fig:mnist}
                \centering
                \includegraphics[width=\linewidth]{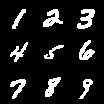}
        \end{subfigure}\hfill
        \begin{subfigure}[b]{0.19\textwidth}
                \caption{MNIST-M}
                \label{fig:mnistm}
                \centering
                \includegraphics[width=\linewidth]{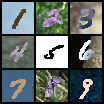}
        \end{subfigure}\hfill
        \begin{subfigure}[b]{0.19\textwidth}
                \caption{MNIST-COCO}
                \label{fig:mnistcoco}
                \centering
                \includegraphics[width=\linewidth]{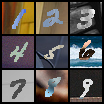}
        \end{subfigure}\hfill
        \begin{subfigure}[b]{0.19\textwidth}
                \caption{PixelDA~\cite{bousmalis_unsupervised}}
                \label{fig:mnist_pda}
                \centering
                \includegraphics[width=\linewidth]{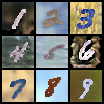}
        \end{subfigure}\hfill
        \begin{subfigure}[b]{0.19\textwidth}
                \caption{Ours}
                \label{fig:mnist_ours}
                \centering
                \includegraphics[width=\linewidth]{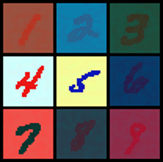}
        \end{subfigure}
        \caption{\textbf{Example samples of the MNIST modalities:} MNIST (Source), MNIST-M (Target), and MNIST-COCO (Generalization) on the left; and example augmentation images generated by PixelDA and our method respectively.}\label{fig:mnist_examples}
\end{figure*}

\subsection{Optimization Objective}
The optimization objective of the deception network is essentially the loss of the recognition network; however, instead of minimizing it, we maximize it by updating the parameters in the direction of the positive gradient. This is achieved by adding a gradient reversal layer \cite{ganin2015unsupervised} between the deception and recognition nets as shown in Fig.~\ref{fig:training}. The layer only negates the gradient when back-propagating, thereby resulting in the reversed optimization objective for a given loss. Therefore, the general optimization objective can be written as follows:
\begin{align}
& \underset{\theta_T}{\text{min}} \hspace{1mm} \underset{\theta_D}{\text{max}}
& & \mathcal{L}_t (T (D (\mathbf{x}; \mathbf{\theta}_D)), \mathbf{y}; \theta_T)\\
& \text{subject to} 
    & & C_{M_n} \hspace{10mm} \text{for  }  \mathbf{n} = 1, \ldots, N_m
\end{align}
where $\mathbf{x}$ is the input image, $\mathbf{y}$ is the ground truth label, $T$ is the task network, $\mathcal{L}_t$ is the task loss, $D$ is the deception network, and $C_m$ denotes the hard constraints defined by the deception modules enforced by projection after a gradient step. In this framework, the deception network's objective only depends on the objective of the recognition task and can, therefore, be easily applied to any other task.


\subsection{Training Procedure}
We use two different SGD solvers, where the actual task network has a learning rate of $0.001$ with a decaying factor of $0.95$ every $20000^{th}$ iteration. The learning rate of the deception network was found to work well with a constant value of $0.01$. We train with a batch size of 64 for all the experiments and we stop training after 500 epochs. 
During the experimentation, we also discovered that concatenating real and perturbed images led to a consistent improvement in numbers. 
  \section{Evaluation}
 \label{sec:exp}

\begin{figure*}[t]
\centering
        \begin{subfigure}[b]{0.19\textwidth}
                \caption{Synthetic}
                \label{fig:lm_s}
                \centering
                \includegraphics[width=\linewidth]{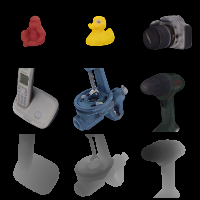}

        \end{subfigure}\hfill
        \begin{subfigure}[b]{0.19\textwidth}
                \caption{Real}
                \label{fig:lm_r}
                \centering
                \includegraphics[width=\linewidth]{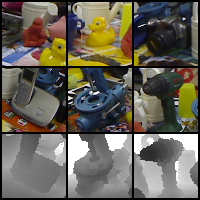}
        \end{subfigure}\hfill
        \begin{subfigure}[b]{0.19\textwidth}
                \caption{Extended}
                \label{fig:lm_e}
                \centering
                \includegraphics[width=\linewidth]{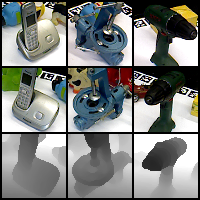}
        \end{subfigure}\hfill
        \begin{subfigure}[b]{0.19\textwidth}
                \caption{PixelDA~\cite{bousmalis_unsupervised}}
                \label{fig:lm_pda}
                \centering
                \includegraphics[width=\linewidth]{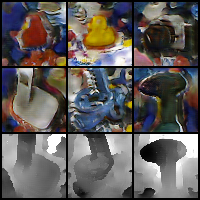}
        \end{subfigure}\hfill
        \begin{subfigure}[b]{0.19\textwidth}
               \caption{Ours}
                \label{fig:lm_ours}
                \centering
                \includegraphics[width=\linewidth]{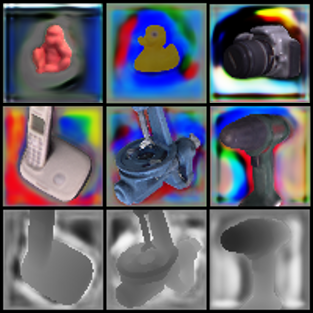}
        \end{subfigure}
        \caption{\textbf{Example samples of the LineMOD modalities:} Synthetic (Source), Real (Target), and Extended (Generalization) on the left; and example augmentation images generated by PixelDA and our method respectively.}\label{fig:lm_examples}
\end{figure*}

In this section, we conduct a series of experiments to compare the capabilities of our pipeline with the state-of-the-art domain adaptation methods. We first compare ourselves against these baselines for the problem of adaptation and will then compare in terms of generalization. We will conclude with an ablative analysis to measure the impact of each module and modality on the final performance.

As the first dataset, we used the popular handwritten digits dataset MNIST as well as MNIST-M, introduced in~\cite{ganin2016domain} for unsupervised domain adaptation (depicted in Figs.~\ref{fig:mnist} ~\ref{fig:mnistm}). MNIST-M blends digits from the original monochrome set with random color patches from BSDS500 \cite{arbelaezContourDetectionHierarchical2010} by simply inverting the color values for the pixels belonging to the digit. The training split containing 59001 target images is then used for domain adaptation. The remaining 9001 target images are used for evaluation. That means that around 86\% of the target data is used for training. Note that while MNIST is not technically synthetic, its clean and homogeneous appearance is typical for synthetic data. 

The second dataset is the Cropped LineMOD dataset ~\cite{Wohlhart15} consisting of small centered, cropped 64$\times$64 patches of 11 different objects in cluttered indoor settings displayed in various of poses. It is based on the LineMOD dataset \cite{hinterstoisser2012model} featuring a collection of annotated RGB-D sequences recorded using the Primesense Carmine sensor and associated 3D object reconstructions. The dataset also features a synthetic set of crops of the same objects in various poses on a black background. We will treat this Synthetic Cropped LineMOD as the source dataset and the Real Cropped LineMOD as the target dataset. Domain adaptation methods use a split of 109208 rendered source images and 9673 real-world target images, 1000 real images for validation, and a target domain test set of 2655 images for testing. We show examples in Figs.~\ref{fig:lm_s} and~\ref{fig:lm_r}.

The last dataset pair we used for the experiments is SYNTHIA~\cite{ros2016synthia} and Cityscapes~\cite{Cordts2016Cityscapes}. SYNTHIA is a collection of pixel-annotated road scene frames rendered from a virtual city. Cityscapes is its real counterpart acquired in the street scenes of 50 different actual cities. Following a common evaluation protocol, we used a subset of 9400 SYNTHIA images, also known as SYNTHIA-RAND-CITYSCAPES, as the source data and 500 Cityscapes validation images as the target data.


\subsection{Adaptation Tests}

All domain adaptation methods use a significant portion of the target data for training, making the resulting mapped source images very similar to the target images (e.g., Fig.~\ref{fig:mnistm} vs \ref{fig:mnist_pda} and Fig.~\ref{fig:lm_r} vs \ref{fig:lm_pda}). 
A common benchmark for domain adaptation is then to compare the performance of a classifier trained on the mapped data against a classifier trained on the source data only (lower baseline) and against a classifier trained directly on the target data (upper baseline).

Our approach is generally disadvantaged since we can structure our domain mapping only through the source data and the deception architecture. To show that our learned randomization is indeed guided, we additionally implement an unguided randomization variant that applies train time augmentation similar to the related work. It employs the same modules and constraints as our deception network, but its perturbations are conditioned on random values in each forward pass instead of latent codes from the input.



\subsubsection{Classification on MNIST}



In Table~\ref{table:baseline} we collect the results of the most relevant methods tested on the MNIST $\rightarrow$ MNIST-M scenario and split them according to the type of data used. Since domain adaptation methods use both source and target data for training, they are allocated to a separate group (\textit{S + T}). Both our method and the unguided randomization variant only have access to the source data and are therefore grouped in \textit{S}. The task network follows the architecture presented in \cite{ganin2015unsupervised}, which is also used by the other methods. The task's objective $\mathcal{L}_t$ is a simple cross entropy loss between the predicted and the ground truth label distributions.

We can identify three key observations: (1) our method shows very competitive results (90.4\% classification) and is on par with the latest domain adaptation pipelines: DSN -- 83.2\%, DRIT -- 91.5\% and PixelDA -- 95.9\%. Moreover, we outperform most of the methods by a significant margin despite the fact that they had access to a large portion of the target data to minimize the domain shift. (2) Guiding the randomization leads to ~7\% higher accuracy which supports our claim convincingly. (3) Surprisingly, unguided randomization (with appropriate modules) alone is in fact enough to outperform most methods on MNIST.

\begin{table}[b]
\resizebox{\columnwidth}{!}{
\begin{tabular}{@{}l|r|c|cc@{}}

\multicolumn{1}{c}{}  &    & \multicolumn{1}{c|}{\shortstack{ MNIST $\rightarrow$ \\ MNIST-M}} & \multicolumn{2}{c}{\shortstack{Synthetic Cropped LineMOD $\rightarrow$ \\ Real Cropped LineMOD}}		\\

\toprule
\textbf{} & \textbf{Model} & \textbf{\shortstack{Classification \\ Accuracy (\%)}} & \textbf{\shortstack{Classification \\ Accuracy (\%)}} & \textbf{\shortstack{Mean \\ Angle Error ($^{\circ}$)}} \\ \midrule
          & Source (S)        & 56.6                                  & 42.9                                  & 73.7                         \\ \midrule
 \multirow{2}{*}{\rotatebox[origin=c]{90}{S}}         & Unguided             &     83.1                                  &       53.1                                &       52.6                       \\
          & Ours           &  90.4                                   &  95.8                                      &  51.9                            \\ \midrule
\multirow{6}{*}{\rotatebox[origin=c]{90}{\shortstack{S + T}}}           & CycleGAN~\cite{zhu2017unpaired}       & 74.5                                  & 68.2                                 & 47.5                        \\
          & MMD~\cite{tzeng2014deep, long2015learning}            & 76.9                                  & 72.4                                 & 70.6                        \\
          & DANN~\cite{ganin2016domain}           & 77.4                                  & 99.9                                  & 56.6                        \\
          & DSN~\cite{bousmalis2016domain}            & 83.2                                  & 100                                   & 53.3                        \\
          & DRIT~\cite{lee2018diverse}             & 91.5                                 & 98.1                                 & 34.4                         \\
          & PixelDA~\cite{bousmalis_unsupervised}        & 95.9                                  & 99.9                                 & 23.5                         \\ \midrule
          & Target (T)         & 96.5                                  & 100                                   & 12.3                         \\ \bottomrule
\end{tabular}
}
\vspace{2pt}
\caption{\textbf{Baseline tests:} While performing slightly worse than the leading state-of-the-art domain adaptation methods using target data, we still manage to achieve very competitive performance without access to target data.}
\label{table:baseline}
\end{table}

\subsubsection{Classification and Pose Estimation on LineMOD}


As before, the domain adaptation methods are trained on a mix of source (Synthetic Cropped LineMOD) and target (Real Cropped LineMOD) data and we compare to the predefined baselines. 
We use the common task network for this benchmark from \cite{ganin2015unsupervised} and the associated task loss:
\begin{equation} \label{eq:task_specific_loss}
\mathcal{L}_{t}(G) = \mathbb{E}_{\mathbf{x^s}, \mathbf{y^s}} \Big[-\mathbf{y^s} ^\top \log{\mathbf{\hat{y}^d}} + \xi \log (1 - \mathbf{q^s} ^\top \mathbf{\hat{q}^d}) \Big]
\end{equation}
where the first term is the classification loss and the second term is the log of a quaternion rotation metric~\cite{huynh2009metrics}. $\xi$ weighs both terms whereas $\mathbf{q^s}$ and $\mathbf{\hat{q}^d}$ are the ground truth and predicted quaternions, respectively. 

The results in Table~\ref{table:baseline} present a more nuanced case. On this visually complex dataset, unguided randomization performs only above the lower baseline and is far behind any other method. Our guided randomization, on the other hand, with -- 95.8\% classification and 51.9$^{\circ}$ angle error is competitive with those of the latest domain adaptation methods using target data: DSN -- 100\% \& 53.3$^{\circ}$, DRIT -- 98.1\% \& 34.4$^{\circ}$, and PixelDA -- 99.9\% \& 23.5$^{\circ}$. Nonetheless, we believe that both DRIT and PixelDA are not fully reachable by target-agnostic methods like ours since the space of all needed adaptations (e.g., aberrations or JPEG artifacts) has to be spanned by our deception modules. The augmentation differences between PixelDA and our method (Figs.~\ref{fig:lm_pda} and ~\ref{fig:lm_ours}) suggest the existance of some visual phenomena we are still not accounting for with our deception network. 

\subsection{Generalization Tests}
For the second set of experiments, we test the generalization capabilities of our method as well as the competing approaches. The major advantage of our pipeline is its independence from any target domain by design. 
To support our case we designed two new datasets: 

\begin{itemize}[itemsep=1mm]
  \item  \textbf{MNIST-COCO} The data collection follows the exact same generation procedure of MNIST-M and has the same exact number of images for both training and testing. The only difference here is that instead of the BSDS500 dataset, we use crops from MS COCO. Fig.~\ref{fig:mnist_ours} demonstrates some of the newly generated images.
  \item  \textbf{Extended Real Cropped LineMOD} Thanks to the help of the authors of the original LineMOD dataset~\cite{hinterstoisser2012model}, we were able to get some of the original LineMOD objects, namely "phone", "benchvise", and "driller". 
We repeated the physical acquisition setup and generated an annotated scene for each object.
Each scene depicts a specific object placed on a white markerboard atop a turntable and coarsely surrounded by a small number of cluttered objects, slightly occluding the object at times. Each sequence contains 130 RGB-D images covering the full 360$^{\circ}$ rotation at an elevation angle of approximately 60$^{\circ}$. Given the acquired and refined poses, we again crop the images in the same fashion as in the Cropped LineMOD dataset~\cite{Wohlhart15}. All 390 images are used for evaluation, with some examples shown in Fig.~\ref{fig:lm_e}. 
\end{itemize}



\begin{table}[b]
\resizebox{\columnwidth}{!}{
\begin{tabular}{@{}l|r|c|cc@{}}

\multicolumn{1}{c}{}  &    & \multicolumn{1}{c|}{\shortstack{MNIST $\rightarrow$ \\ MNIST-COCO}} & \multicolumn{2}{c}{\shortstack{Synthetic Cropped LineMOD $\rightarrow$ \\ Extended Real Cropped LineMOD}}		\\

\toprule
\textbf{} & \textbf{Model} & \textbf{\shortstack{Classification \\ Accuracy (\%)}} & \textbf{\shortstack{Classification \\ Accuracy (\%)}} & \textbf{\shortstack{Mean \\ Angle Error ($^{\circ}$)}} \\ \midrule
          & Source (S)        & 57.2                                  & 63.1                                  & 78.3                         \\ \midrule
 \multirow{2}{*}{\rotatebox[origin=c]{90}{S}}         & Unguided             &     85.8                                  &       77.2                                &       48.5                       \\
          & Ours           &  89.4                                   &  99.0                                      &  46.5                           \\ \midrule
\multirow{2}{*}{\rotatebox[origin=c]{90}{\shortstack{S + T}}}  
          & DSN~\cite{bousmalis2016domain}             & 73.2                                  & 45.7                                   & 76.3              \\
          & PixelDA~\cite{bousmalis_unsupervised}        &    72.5                               & 76.0                                 & 84.2                         \\ \midrule
          & Target (T)         & 96.1                                  & 100                                   & 14.7                         \\ \bottomrule
\end{tabular}
}
\caption{\textbf{Generalization tests:} Our method generalizes well to the extended datasets, while the adaptation methods underperform due to overfitting.}
\label{table:generalization}
\end{table}

For a comparison with the strongest related methods, i.e., DSN, DRIT, and PixelDA, we used open source implementations and diligently ensured that we are able to properly train and reproduce the reported numbers from Table~\ref{table:baseline}. While the DRIT implementation worked well for the adaptation experiments, we failed to produce reasonably high numbers for the generalization experiment and chose to exclude it from the comparison.


Similar to before, we train them using the target data from MNIST-M and Real Cropped LineMOD. After the training is finished and the corresponding accuracies on the target test splits are achieved, we test them on the newly acquired dataset. While different, these extended datasets still bear a certain resemblance to the target dataset and we could expect to see a certain amount of generalization. 
For our randomization methods, we can immediately test on the new data, since retraining is not necessary.


\begin{table}[b]
\resizebox{\columnwidth}{!}{
    \begin{tabular}{r|c|cc}
      &  \multicolumn{1}{c|}{\shortstack{MNIST $\rightarrow$ \\ MNIST-M}} & \multicolumn{2}{c}{\shortstack{Synthetic Cropped LineMOD $\rightarrow$ \\ Real Cropped LineMOD}} \\
    \midrule
    \textbf{Modules} & \textbf{\shortstack{Classification \\ Accuracy (\%)}} & \textbf{\shortstack{Classification \\ Accuracy (\%)}} & \textbf{\shortstack{Mean \\ Angle Error ($^{\circ}$)}} \\ 
    \midrule
    None & 56.6 & 42.9 & 73.7 \\
    \midrule
    BG & 82.4 & 74.8 & 50.4 \\
    BG + NS & 86.5 & 77.6 & 52.8 \\
    BG + NS + DS & 90.4 & 78.7 & 48.2 \\
    BG + NS + DS + L & - & 95.8 & 51.9 \\
    \bottomrule
    \end{tabular}%
}
\caption{\textbf{Module ablation:} Evaluation of the importance of the deception network's modules. BG -- background, NS -- noise, DS -- distortion, L -- light.}
\label{table:modules}
\end{table}

\setlength{\tabcolsep}{8pt}
\begin{table*}[t]
  \centering
  \resizebox{1\textwidth}{!}{%
    \begin{tabular}{r|r|cccccccccccccccc|c|c}
    \multicolumn{1}{r}{} &   & Road & SW & BLDG & Wall & Fence & Pole & TL & TS & VEG & Sky & PRSN & Rider & Car & Bus & Mbike & Bike & \textbf{mIoU} & \textbf{mIoU*} \\
    \midrule
      & Source (S) & 3.8 & 10.2 & 46.3 & 1.8 & 0.3 & 19.1 & 4.0 & 7.5 & 71.8 & 72.2 & 44.6 & 3.4 & 24.9 & 5.2 & 0.0 & 2.5 & 19.8 & 22.8 \\
    \midrule
    \multicolumn{1}{c|}{\multirow{2}[2]{*}{\begin{sideways}S\end{sideways}}} & Unguided & 17.9 & 8.8 & 59.2 & 0.8 & 0.4 & 22.1 & 3.5 & 6.1 & 71.4 & 70.4 & 40.3 & 7.3 & 37.9 & 3.3 & 0.2 & 7.3 & 22.3 & 25.7 \\
      & Ours & 51.4 & 17.8 & 62.5 & 1.6 & 0.4 & 22.6 & 6.0 & 11.9 & 70.9 & 73.5 & 42.1 & 8.2 & 40.9 & 8.1 & 3.9 & 18.4 & 27.5 & 32.0 \\
    \midrule
    \multicolumn{1}{c|}{\multirow{7}[2]{*}{\begin{sideways}S + T\end{sideways}}} & FCNs Wld~\cite{hoffman2016fcns} & 11.5 & 19.6 & 30.8 & 4.4 & 0.0 & 20.3 & 0.1 & 11.7 & 42.3 & 68.7 & 51.2 & 3.8 & 54.0 & 3.2 & 0.2 & 0.6 & 20.1 & 22.9 \\
      & CDA~\cite{zhang2017curriculum} & 65.2 & 26.1 & 74.9 & 0.1 & 0.5 & 10.7 & 3.7 & 3.0 & 76.1 & 70.6 & 47.1 & 8.2 & 43.2 & 20.7 & 0.7 & 13.1 & 29.0 & 34.8 \\
      & Cross-City~\cite{chen2017no} & 62.7 & 25.6 & 78.3 & - & - & - & 1.2 & 5.4 & 81.3 & 81.0 & 37.4 & 6.4 & 63.5 & 16.1 & 1.2 & 4.6 & - & 35.7 \\
      & Tsai et al.~\cite{tsai2018learning}& 78.9 & 29.2 & 75.5 & - & - & - & 0.1 & 4.8 & 72.6 & 76.7 & 43.4 & 8.8 & 71.1 & 16.0 & 3.6 & 8.4 & - & 37.6 \\
      & ROAD-Net~\cite{chen2018road} & 77.7 & 30.0 & 77.5 & 9.6 & 0.3 & 25.8 & 10.3 & 15.6 & 77.6 & 79.8 & 44.5 & 16.6 & 67.8 & 14.5 & 7.0 & 23.8 & 36.1 & 41.7 \\
      & LSD-seg~\cite{sankaranarayanan2018learning} & 80.1 & 29.1 & 77.5 & 2.8 & 0.4 & 26.8 & 11.1 & 18.0 & 78.1 & 76.7 & 48.2 & 15.2 & 70.5 & 17.4 & 8.7 & 16.7 & 36.1 & 42.1 \\
      & Chen et al.~\cite{chen2019learning} & 78.3 & 29.2 & 76.9 & 11.4 & 0.3 & 26.5 & 10.8 & 17.2 & 81.7 & 81.9 & 45.8 & 15.4 & 68.0 & 15.9 & 7.5 & 30.4 & 37.3 & 43.0 \\
    \midrule
      & Target (T) & 96.5 & 74.6 & 86.1 & 37.1 & 33.2 & 30.2 & 39.7 & 51.6 & 87.3 & 90.4 & 60.1 & 31.7 & 88.4 & 52.3 & 33.6 & 59.1 & 59.5 & 65.5 \\
    \bottomrule
    \end{tabular}%
  }
  \caption{\textbf{Real-world application:} Segmentation performance on SYNTHIA $\rightarrow$ Cityscapes benchmark based on Intersection over Union (IoU) tested on 16 (\textbf{mIoU}) and 13 (\textbf{mIoU*}) classes of the Cityscapes dataset. Our method outperforms \textit{source} and \textit{unguided} by a significant margin and remains competitive to the methods relying on the target data.}
  \vspace{-8pt}
\label{table:cars}%
\end{table*}%

As is evident from Table \ref{table:generalization}, the accuracy of our method on MNIST-COCO is very close to the MNIST-M number (90.4\% and 89.4\% respectively). For the case of Extended Real Cropped LineMOD, we get even better results than for the Real Cropped LineMOD for both accuracy and angle error: We only need to classify 3 objects instead of 11 with a much smaller pose space, and the scenes are in general cleaner and less occluded. These results underline our claim with respect to generalization. This is, however, not the case for the domain adaptation methods showing drastically worse results. Interestingly, we observe an inverse trend where better results on the original target data lead to a more significant drop. Despite of having a very high accuracy on the target data and the ability to generate additional samples that do not exist in the dataset, these methods present typical signs of overfit mappings that cannot generalize well to the extensions of the same data acquired in a similar manner. The simple reason for this might be the nature of these methods:
they do not generalize to the features that matter the most for the recognition task, but to simply replicate the target distribution as close as possible. As a result, it is not clear what the classifier exactly focuses on during inference; however, it could very likely be the particular type of images (e.g., in case of MNIST-COCO) or a specific type of backgrounds and illumination (e.g., in case of Extended Real Cropped LineMOD). In contrast to domain adaptation methods, our pipeline is designed not to replicate the target distribution, but to make the classifier invariant to the changes that should not affect classification, which is the reason why our results remain stable.

\subsection{Ablation Studies}
In this section, we perform a set of ablation studies to gain more insight into the impact of each module inside the deception network. Obviously, our modules model only a fraction of possible perturbations and it is important to understand the individual contributions. Moreover, we demonstrate how well we perform provided different types of input modalities for the LineMOD datasets.

\begin{table}[b]
\vspace{-0.2em}
\resizebox{\columnwidth}{!}{
    \begin{tabular}{r|cc|cc}
      & \multicolumn{2}{c|}{\shortstack{Synthetic Cropped LineMOD $\rightarrow$ \\ Real Cropped LineMOD}} & \multicolumn{2}{c}{\shortstack{Synthetic Cropped LineMOD $\rightarrow$ \\ Extended Real Cropped LineMOD }} \\
    \midrule
    \textbf{Input} & \textbf{\shortstack{Classification \\ Accuracy (\%)}} & \textbf{\shortstack{Mean \\ Angle Error ($^{\circ}$)}} & \textbf{\shortstack{Classification \\ Accuracy (\%)}} & \textbf{\shortstack{Mean \\ Angle Error ($^{\circ}$)}} \\
    \midrule
    D & 73.3 & 36.6 & 78.7 & 34.9 \\
    RGB & 84.8 & 57.4 & 85.9 & 49.4 \\
    RGB-D & 95.8 & 51.9 & 99.0 & 46.5 \\
    \bottomrule
    \end{tabular}%

}
\caption{\textbf{Input modality ablation:} Performance evaluation based on the input data type used: depth, RGB, or RGB-D.}
\label{table:modalities}
\end{table}

\subsubsection{Deception Modules}
We tested 4 different variations of the deception net that use varying combinations of the deception modules: \textit{background (BG)}, \textit{noise (NS)}, \textit{distortion (DS)}, and \textit{light (L)}. The exact combinations and the results on both datasets are listed in Table~\ref{table:modules}.

It can be clearly seen that each additional module in the deception network adds to the discriminative power of the final task network. The most important modules can also be easily distinguished based on the results. Apparently, the background module always makes a significant difference: the purely black backgrounds of the source data are drastically different from the real imagery. Another interesting observation is the strong impact the lighting perturbation has in the case of the Cropped LineMOD dataset. This enforces the notion that real sequences undergo many kinds of lighting changes that are not well-represented by synthetic renderings without any additional relighting. Note that the MNIST deception network does not employ lighting.

\subsubsection{Input Modalities}
For the task of simultaneous instance classification and pose estimation, we (as well as the other methods) always use the full RGB-D information. 
This ablation aims to show how well we fare provided only a certain type of data and the impact on the final results. Table~\ref{table:modalities} shows that RGB allows for better classification, whereas depth provides better pose estimates. We can further boost the classification enormously and reduce the pose error by combining both modalities.

\subsection{Real-world Scenario}

We demonstrate a real-world application of our approach on a more practical problem of semantic segmentation using the common SYNTHIA $\rightarrow$ Cityscapes benchmark. Having only synthetic SYNTHIA renderings, we try to generalize to the real Cityscapes data by evaluating our method on 13 and 16 classes using the Intersection over Union (IoU) metric. This setup is particularly difficult since the domain gap problem here is intensified by a completely different set of segmentation instances and camera views. For a fair comparison, all methods use a VGG-16 base (FCN-8s) recognition network. The deception modules used in this case are as follows: 2D noise (NS), elastic distortion (DS), and light (L). Normal maps for the light module are generated from the available synthetic depth data.


Table~\ref{table:cars} shows that even without access to target domain data, our pipeline remains competitive with the methods relying on target data, showing mIoU of 27.5$\%$ and mIoU* of 32$\%$ (16 and 13 classes) -- well above \textit{source} and \textit{unguided}. The results also confirm the generality of the approach with respect to the different task architectures and datasets.

 \section{Conclusion}
 \label{sec:cnc}

In this paper we presented a new framework to tackle the domain gap problem when no target data is available. Using a task network and its objective, we show how to extend it with a simple encoder-decoder deception network and bind both in a min-max game in order to achieve guided domain randomization. As a result, we obtain increasingly more robust task networks. We demonstrate a comparable performance to domain adaptation methods on two datasets and, most importantly, show superior generalization capabilities where the domain adaptation methods tend to drop in performance due to overfitting to the target distribution.
Our results suggest that guided randomization, because of its simple but effective nature, should become a standard procedure to define baselines for domain transfer and adaptation techniques.

 \vfill
 
 \begin{figure*}[h]
	\centering
	\includegraphics[width=.98\linewidth]{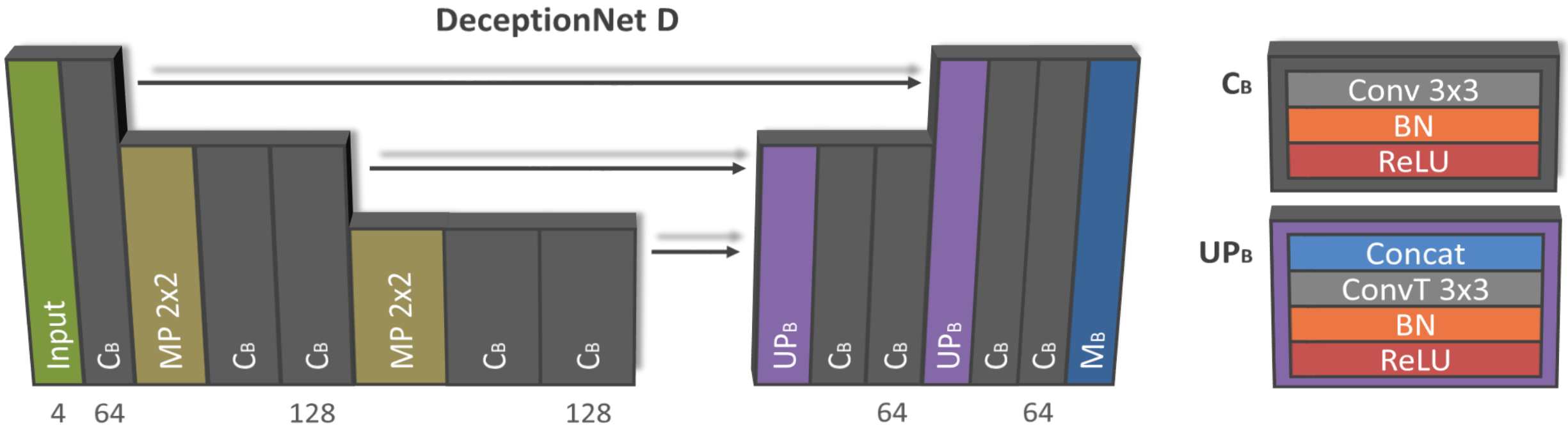}
	\caption{\textbf{DeceptionNet architecture}. Our network features a typical encoder-decoder architecture. The encoder part consists of 2 consecutive downsamplings followed by a sequence of convolutional blocks $C_B$. The decoder part shares a similar architecture for all presented augmentation modules. Arrows show the skip connections between blocks.}
	\label{fig:deceptionnet}  
\end{figure*}

\appendix
 \section{Supplementary Material}
 \label{sec:cnc}
 \subsection{Network Architecture}

Let $C^B_k$ be a convolutional block composed of the following layers: 3$\times$3 convolution with $k$ filters, BatchNorm (BN), and ReLU activation function. Similarly, let ${UP}_B$ be a decoding block made of: 2-factor upsampling transposed convolution, BatchNorm (BN), and ReLU.

\paragraph{DeceptionNet $\mathbf{D}$:} Using the defined nomenclature, the encoding part of the DeceptionNet can be described as: $C^B_{64}-MP-C^B_{128}-C^B_{128}-MP-C^B_{128}-C^B_{128}$; and its decoding part as: $UP_{64}-C^B_{64}-C^B_{64}-UP_{64}-C^B_{64}-C^B_{64}-M_B$. Where $M_B$ is defined by the specific module type, and $MP$ stands for a 2-factor max-pooling layer. Encoding blocks have skip connections concatenating the channels with the opposite decoding blocks. The visual representation of the DeceptionNet's architecture is depicted in Fig.~\ref{fig:deceptionnet}.
	
\begin{figure}[b]
	\centering
	\includegraphics[width=0.9\linewidth]{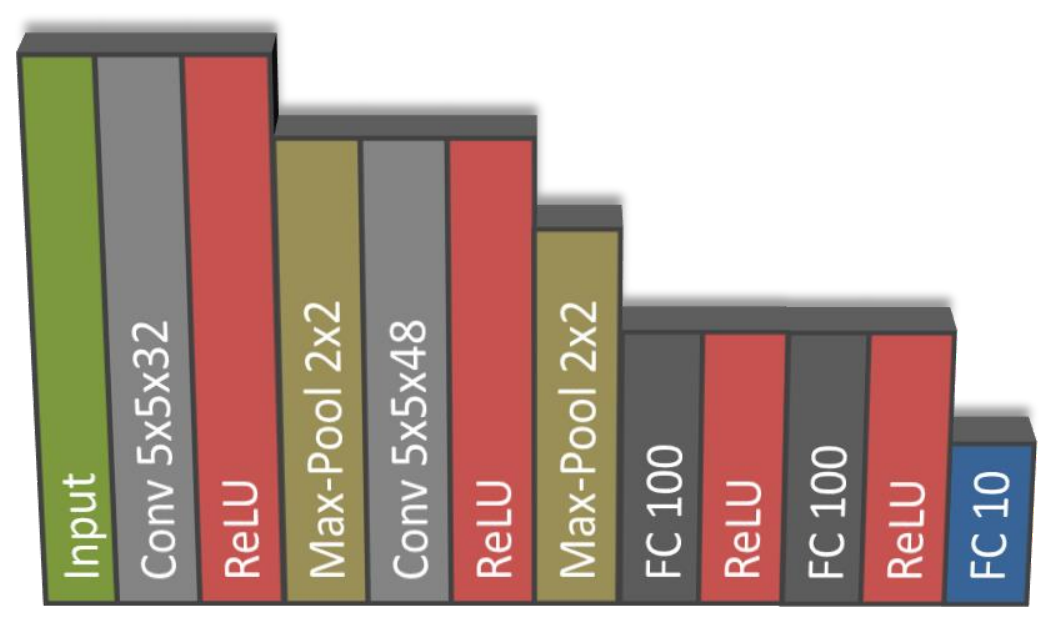}
	\caption{\textbf{MNIST Classifier:} Simple LeNet-like architecture, where 2 convolutional layers followed by ReLUs and max-poolings are finalized by 3 fully-connected layers.}
	\label{fig:ic}
\end{figure}
	
\begin{figure}[b]
	\centering
	\includegraphics[width=0.9\linewidth]{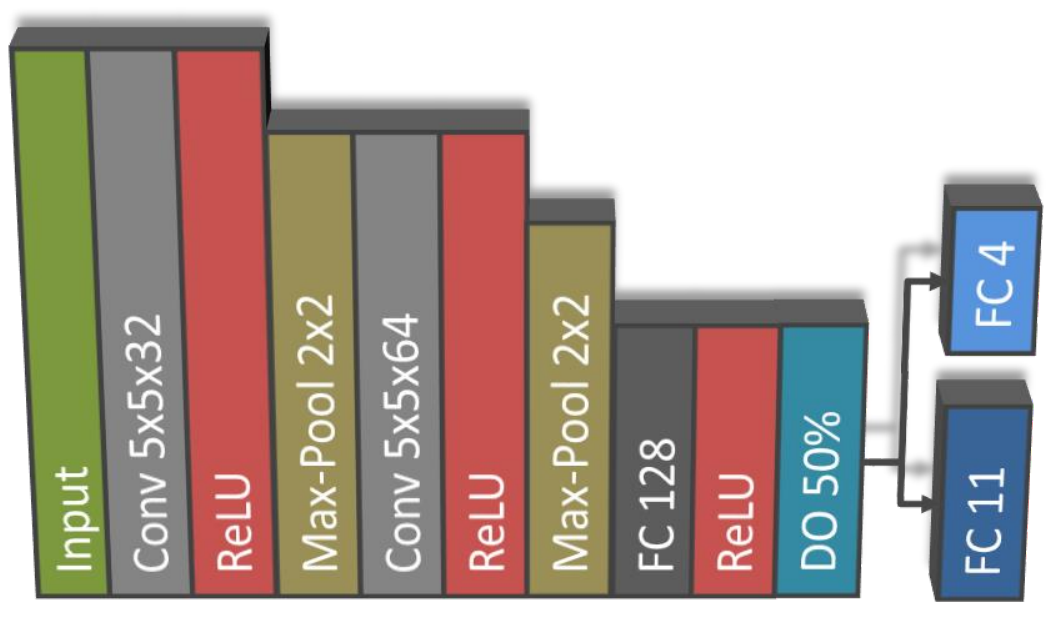}
	\caption{\textbf{Cropped LineMOD Task Network:} Simple LeNet-like architecture followed by a dropout layer with a 50\% rate and outputting both a class and pose vector.}
	\label{fig:icpe}
\end{figure}

\paragraph{Task Network $\mathbf{T}$:} In both cases, i.e., for MNIST classification and Cropped LineMOD classification and pose estimation, $T$ follows a simple LeNet-like architecture. As for MNIST (see Fig.~\ref{fig:ic}), the final layer outputs the 10D vector, whereas for Cropped LineMOD (see Fig.~\ref{fig:icpe}) there is a 11D classification output as well as 4D quaternion output.

\begin{figure*}[b]
	\centering
	\includegraphics[width=1\textwidth]{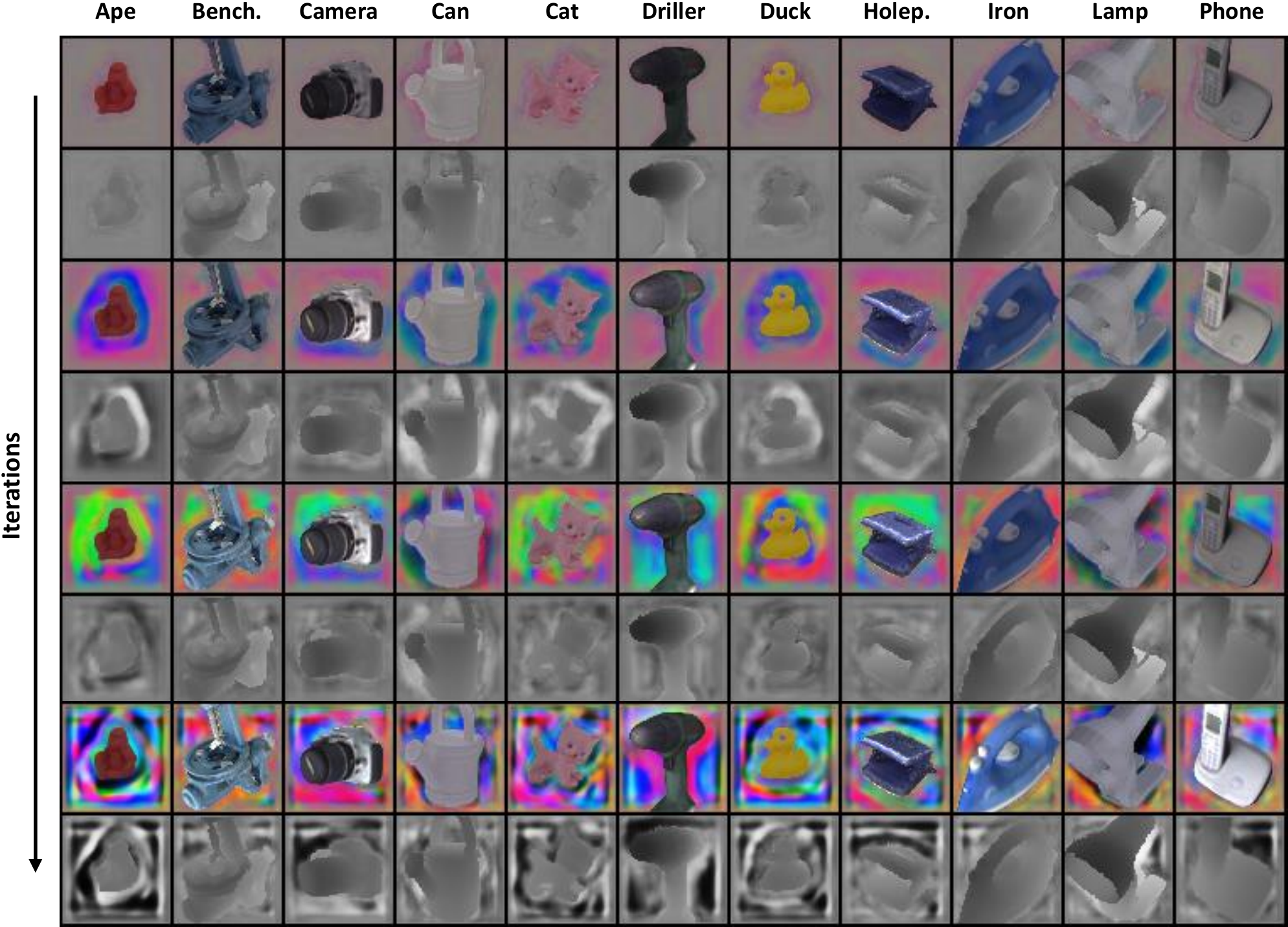}
	\caption{\textbf{Deceptive images $\mathbf{x^d}$ over consecutive iterations:} The output becomes increasingly more complex for $T$.}
	\label{fig:addimages}  
\end{figure*}

\begin{figure}[h]
	\centering
	\includegraphics[width=1\linewidth]{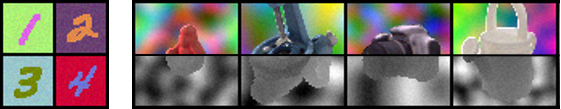}
	\caption{\textbf{Unguided samples:} We provide a sample of unguided augmentations for MNIST and LineMOD.}
	\label{fig:dr}
\end{figure}

\begin{figure}[h]
	\centering
	\includegraphics[width=1\linewidth]{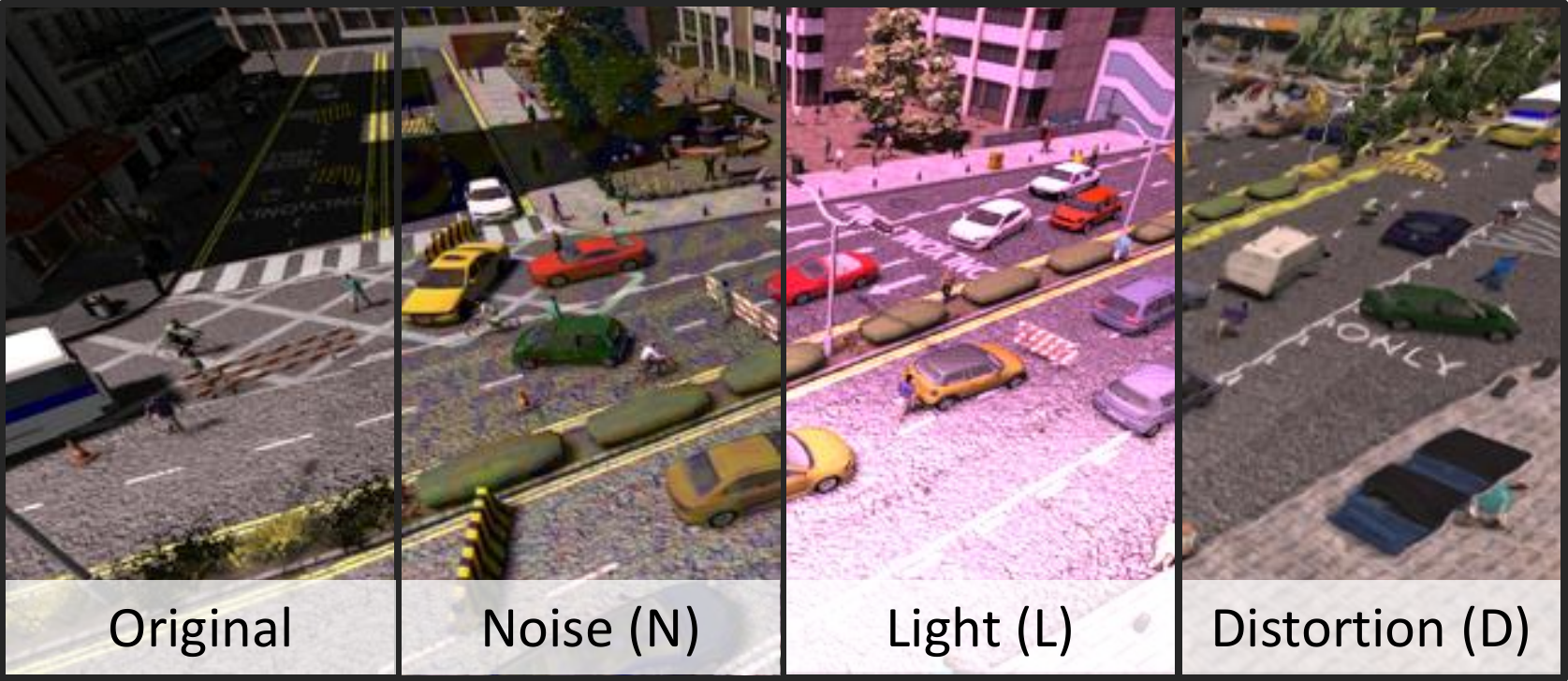}
    \caption{\textbf{Deceptive augmentations:} Augmentations applied for the SYNTHIA $\rightarrow$ Cityscapes scenario.}
	\label{fig:cars}  
\end{figure}

\subsection{Unguided Randomization: BG Filling}
One of the modalities we have compared our results with is unguided randomization that applies augmentations during the data preprocessing step. While using the same modules and constraints as our deception network, its perturbations are conditioned on random values instead of latent codes from the input.
    
Since our DeceptionNet is capable of generating very complex backgrounds, we have also used complex noise types for unguided randomization to make the comparison more fair (see Fig.~\ref{fig:dr}). Apart from a uniform white noise, two additional noise types were used: Perlin ~\cite{perlin2002improving} and cellular noises~\cite{worley1996cellular}. Sample frequencies were sampled from the uniform distribution $[0.0001,0.1]$. Both noise types were generated using the open source FastNoise library~\cite{fastnoise}. 
\subsection{Additional Qualitative Results}
In this section, we present additional output examples of the deception networks for Synthetic Cropped LineMOD and SYNTHIA test cases. 

The LineMOD deception network uses all of the deception modules presented in the paper, whereas the SYNTHIA deception network uses three modules: light (L), elastic distortions (DS), and foreground noise (N). The sample outputs from each of the above-mentioned modules are shown in Fig.~\ref{fig:cars}. Moreover, Fig.~\ref{fig:addimages} demonstrates the output of the deception network during the training process. One can see that the output becomes increasingly more sophisticated for recognition by the task network.

\vfill

{\small
\bibliographystyle{ieee_fullname}
\bibliography{egbib}
}

\end{document}